%% file: main.tex
\documentclass[letterpaper, 10 pt, conference]{ieeeconf}  

\IEEEoverridecommandlockouts                              

\input{packages}

\input{macros}

\title{\huge \bf \OURS: A Multimodal Tactile Finger with\\Taxelized Dynamic Sensing for Dexterous Manipulation}

\author{
Eric T. Chang\authorrefmark{1}$^1$,
Peter Ballentine\authorrefmark{1}$^2$,
Zhanpeng He\authorrefmark{1}$^3$,
Do-Gon Kim$^1$,
Kai Jiang$^3$,
Hua-Hsuan Liang$^3$,\\ %
Joaquin Palacios$^1$,
William Wang$^3$,
Pedro Piacenza$^1$,
Ioannis Kymissis$^2$, %
Matei Ciocarlie$^1$ \\ %
\thanks{}  
}

\begin{document}

\twocolumn[{%
\renewcommand\twocolumn[1][]{#1}%
\maketitle
\begin{center}
    \centering
    \vspace{-3em}
    \captionsetup{type=figure, font=small}
    \includegraphics[width=1.0\textwidth]{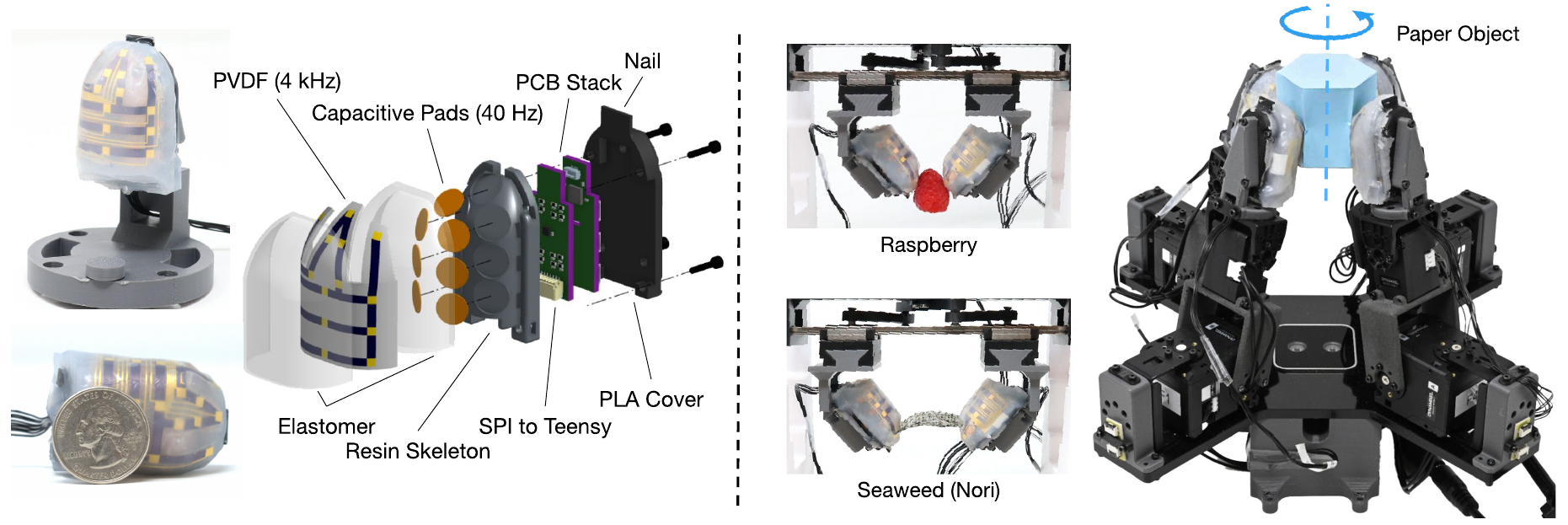}
    \captionof{figure}{\textbf{Overview of \OURS}. Left: completed finger and design. Right: integration on a parallel gripper and multifingered robot hand. Combining high-performance, taxelized PVDF for dynamic sensing with capacitive pads for static sensing, \OURS enables fast yet delicate manipulation. Used in conjunction with imitation learning and on-robot reinforcement learning fine-tuning, \OURS data can also enable multifingered dexterity such as in-hand reorientation even on fragile objects.}
    \label{fig:candy}
\end{center}
}]
{
  \renewcommand{\thefootnote}%
    {\fnsymbol{footnote}}
  \footnotetext[1]{indicates joint-first authorship.}
  \footnotetext[0]{$^1$ Dept. of Mechanical Engineering $^2$ {Dept. of Electrical Engineering $^3$ Dept. of Computer Science, Columbia University, New York, NY 10027, USA. \hspace{.3cm}This work was supported by a NASA Space Technology Graduate Research Opportunity and by a National Science Foundation Graduate Research Fellowship under Grant No. DGE-2036197. Correspondence email: {\tt\small eric.chang@columbia.edu}}
  }

\thispagestyle{empty}
\pagestyle{empty}

\begin{abstract}
\input{text/abstract.tex}
\end{abstract}
\input{text/intro.tex}
\input{text/related.tex}

\input{text/design}
\input{text/characterization}

\input{text/rh3}
\input{text/discussion}

\input{text/conclusion}

\input{text/ack}

\bibliographystyle{IEEEtran}
\bibliography{
    bib/IEEEabrv,
    bib/myabrv,
    bib/mylibrary,
    bib/other,
    bib/rl
    }

\addtolength{\textheight}{-12cm}   
\end{document}

%% file: packages.tex
\usepackage{color}
\usepackage{epsfig}
\usepackage{graphicx}
\usepackage{algorithm,algorithmic}

\usepackage{adjustbox}
\usepackage{array}
\usepackage{booktabs}
\usepackage{colortbl}
\usepackage{float,wrapfig}
\usepackage{framed}
\usepackage{hhline}
\usepackage{multirow}
\usepackage{makecell}
\usepackage[percent]{overpic}

\usepackage{amsmath,amsfonts,amssymb}
\usepackage{amsthm} 
\usepackage{bm}
\usepackage{nicefrac}
\usepackage{microtype}
\usepackage{contour}
\usepackage{courier}
\usepackage{siunitx}

\usepackage{changepage}
\usepackage{extramarks}
\usepackage{fancyhdr}
\usepackage{lastpage}
\usepackage{setspace}
\usepackage{soul}
\usepackage{xspace}
\usepackage{cuted}
\usepackage{fancybox}
\usepackage{afterpage}
\usepackage{gensymb}

\usepackage[breaklinks=true,colorlinks,backref=True]{hyperref}
\hypersetup{colorlinks,linkcolor={black},citecolor={MSBlue},urlcolor={magenta}}
\usepackage{url}
\usepackage{quoting}
\usepackage{epigraph}

\usepackage{enumerate}
\usepackage{paralist,tabularx}
\usepackage{comment}
\usepackage{pdfpages}
\usepackage{caption}  

\usepackage{pifont}

\usepackage{MnSymbol}

\usepackage{bbold}

\usepackage[para]{footmisc}
\usepackage{cite}

%% file: macros.tex
\makeatletter
\DeclareRobustCommand\onedot{\futurelet\@let@token\@onedot}
\def\@onedot{\ifx\@let@token.\else.\null\fi\xspace}

\makeatother

\definecolor{MyDarkBlue}{rgb}{0,0.08,1}
\definecolor{MyDarkGreen}{rgb}{0.02,0.6,0.02}
\definecolor{MyDarkRed}{rgb}{0.8,0.02,0.02}
\definecolor{MyDarkOrange}{rgb}{0.40,0.2,0.02}
\definecolor{MyPurple}{RGB}{111,0,255}
\definecolor{MyRed}{rgb}{1.0,0.0,0.0}
\definecolor{MyGold}{rgb}{0.75,0.6,0.12}
\definecolor{MyDarkgray}{rgb}{0.66, 0.66, 0.66}
\definecolor{MyPink}{rgb}{1, 0.75, 0.79}
\definecolor{GreenStarColor}{rgb}{0.54, 0.84, 0.41}
\definecolor{MSBlue}{rgb}{0, 0.35, 0.49}

\def\OURS{SpikeATac\xspace}



%% file: text/abstract.tex
In this work, we introduce SpikeATac, a multimodal tactile finger combining a taxelized and highly sensitive dynamic response (PVDF) with a static transduction method (capacitive) for multimodal touch sensing. Named for its `spiky' response, SpikeATac's 16-taxel PVDF film sampled at 4 kHz provides fast, sensitive dynamic signals to the very onset and breaking of contact. We characterize the sensitivity of the different modalities, and show that SpikeATac provides the ability to stop quickly and delicately when grasping fragile, deformable objects. Beyond parallel grasping, we show that SpikeATac can be used in a learning-based framework to achieve new capabilities on a dexterous multifingered robot hand. We use reinforcement learning from human feedback to fine-tune the behavior of a policy to modulate force. Our hardware platform and learning pipeline together enable a difficult dexterous and contact-rich task that has not previously been achieved: in-hand manipulation of fragile objects. Videos are available at \href{https://roamlab.github.io/spikeatac/}{roamlab.github.io/spikeatac}.

%% file: text/intro.tex
\section{Introduction}
\label{sec:intro}

Tactile sensing typically falls into two categories: static sensing, for measuring sustained or slowly varying pressure, often with high \textit{spatial resolution}, and dynamic sensing, for detecting rapid changes in pressure with high \textit{temporal resolution}. These modes mirror biological mechanoreceptors, with the human skin's SA-I/II responding to sustained pressure and stretch and FA-I/II to dynamic changes.

These types of sensing are highly complementary. While static sensors can tell us about contact forces, contact geometry, and surface features, dynamic tactile sensors can reveal the precise moment when contact is made or broken, capture slip events, and measure vibrations created from extrinsic contacts and surface textures \cite{johansson_coding_2009, cutkosky_dynamic_2014}. Both are an important part of rich, multimodal tactile sensation.

Given the promise of dynamic sensing, the field has dedicated important resources to transduction methods that can convert this promise into advanced manipulation abilities. In particular, polyvinylidene fluoride (PVDF) has emerged as a particularly promising material for dynamic sensing. It is highly sensitive to dynamic events, offers high frequency bandwidth, is low cost, and can be manufactured in thin, flexible, and customizable form factors with custom taxel arrangements \cite{kappassov_tactile_2015, bae_pvdf-based_2019, yu_tri-modal_2018, hsu_locally_2011}. Furthermore, amplification circuits can even be fabricated directly on the PVDF film \cite{dahiya_design_2009, hsu_locally_2011}, giving it a high ceiling for scaling to larger and integrated arrays.

In principle, these properties make PVDF well suited for creating dense, flexible, dynamic arrays. However, PVDF has almost exclusively been implemented in robot fingers as large, off-the-shelf taxels providing large-coverage, single-taxel sensing. Challenges remain in (a) how to build taxelized PVDF sheets that can integrate into a finger, (b) how to complement PVDF with a static modality, and (c) how to integrate its rich, difficult-to-simulate signals into modern, data-driven manipulation pipelines.

In this work, we address all three challenges with \textbf{SpikeATac}, a multimodal tactile finger containing a high frequency (4 kHz), 16-taxel PVDF array embedded at the contact surface, complemented by 7 capacitive pads embedded below the PVDF for pressure sensing. An optional accelerometer inside the finger provides additional high frequency feedback. At $45\times32\times25$ mm (length $\times$ width $\times$ thickness), SpikeATac is just larger than a human thumb, has a finger-shaped form factor conducive to stable contact states, and has \SI{180}{\degree} sensing coverage.

We demonstrate the utility of this new design in multiple ways. First, we show that SpikeATac enables rapid yet delicate manipulation, even of fragile objects, by leveraging the extreme responsiveness of PVDF to contact onset. Second, we integrate SpikeATacs into a four-finger robotic hand and show that raw SpikeATac signals can be incorporated into an imitation and reinforcement learning fine-tuning pipeline. We provide a recipe for learning on real, complex SpikeATac signals which achieves high dexterity even on fragile and delicate objects. Overall, our primary contributions are:

\begin{itemize}
\item To our knowledge, we are the first to propose a method for creating a robotic finger containing PVDF arrays with many individuated taxels across the entire surface. This design allows us to combine multi-taxel, dynamic, high frequency sensing with an additional transduction method (capacitive in our case) for static pressure. 
\item We leverage the extreme dynamic sensitivity of PVDF by complementing it with static sensing to achieve high-performance manipulation. We show an ability to perform very fast yet delicate manipulation of extremely fragile objects, an ability that, to our knowledge, has not been demonstrated before.
\item We also show that highly sensitive, dynamic tactile sensing can be integrated into modern, learning-based motor policies. Specifically, we use reinforcement learning from human feedback (RLHF) to fine-tune the behaviors of a policy to precisely modulate force. We demonstrate this via a multimodal policy combining dynamic and static sensing to achieve dexterous, multifingered manipulation of fragile objects, an ability that, to our knowledge, has not been shown before.
\end{itemize}

%% file: text/related.tex
\section{Related Work}
\label{sec:related}

\subsubsection{Dynamic Sensing}

Examples of dynamic modalities for measuring surface vibration include event-based piezoresistive arrays \cite{taunyazov_extended_2021}, capacitive sensors instrumented dynamically \cite{cutkosky_dynamic_2014}, hydrophones \cite{fishel_bayesian_2012}, MEMS microphones \cite{pestell_artificial_2022, andrussow_adding_2025, lambeta_digitizing_2024, chang_investigation_2024, yang_touch_2023}, and accelerometers or piezoelectrics embedded in the finger surface \cite{howe_dynamic_1993, son_tactile_1994}.
We believe that high vibration signal dimensionality and spatial distribution are valuable, especially to a learned model or policy which has the ability to parse complex, overlapping signals and time-series data \cite{taunyazov_extended_2021, piacenza_sensorized_2020, yang_touch_2023}.
However, scaling these sensors to high-dimensional arrays is nontrivial due in part to transducer form factors, wiring, and sampling challenges.

\subsubsection{PVDF-based Fingertips}
PVDF is an established dynamic sensing modality in robotics, and previous work has shown its capability to sense contact events, slip events, and textures \cite{cutkosky_dynamic_2014, kappassov_tactile_2015, son_tactile_1994, howe_dynamic_1993, howe_grasping_1990, shi_surface_2023, guo_vibrotact_2024}.
A number of works have combined PVDF strips with a static sensing modality in a fingertip \cite{Choi_development_2006, shi_surface_2023, Goger_tactile_2009, ke_fingertip_2019}, however these works utilize only one or two PVDF strips, limiting their resolution.
One study combined multi-taxel PVDF with a piezoresistive array \cite{yu_tri-modal_2018}, and another demonstrated how PVDF can be used for shear sensing when fabricated in dense multi-taxel modules \cite{yu_flexible_2016}.
Neither of these works, however, demonstrate multi-taxel PVDF in a fingertip.

\subsubsection{Multimodal Fingers}

Beyond PVDF, one can find many multimodal finger designs in the literature using a variety of sensing modalities. While many are flat form factors \cite{lu_gtac_2022, lima_dynamic_2020, martinez-hernandez_soft_2023, romano_human-inspired_2011}, we focus on multi-curved fingers that can sense diverse contacts during dexterous manipulation. Saloutos et al. built a spherical multimodal fingertip combining barometers and time-of-flight sensors \cite{saloutos_design_2023}, BioTac combines impedance sensing electrodes in both DC and AC configurations, temperature sensors, and a hydrophone for higher frequency vibration sensing \cite{fishel_bayesian_2012}, and TacTip and Minsound add a MEMS microphone to a camera-based tactile finger \cite{pestell_artificial_2022, andrussow_adding_2025}. Digit360 combines many modalities into a compact fingertip, including a camera, IMU, MEMS microphones, barometers, temperature sensors, and gas sensors \cite{lambeta_digitizing_2024}.

All of these are valuable platforms for multimodal, dynamic manipulation. \textit{The primary innovation of our work is the exploration of multi-taxel PVDF in a multimodal finger.} Multi-taxel PVDF provides distributed sensing across the finger, rather than only a few dynamic, high frequency ($>$ 1 kHz) transducers.
Lastly, we note that high bandwidth static sensors can also capture high frequency information.
We favor a multimodal approach with designated dynamic sensors like PVDF so that the dynamic modality can be tuned to be sensitive to small transient events, while static sensors can preserve a larger dynamic range.

\subsubsection{On-robot RL fine-tuning}
RL has proven to be an effective approach for policy adaptation and improvement.
A growing line of research \cite{smith2024grow, smith2022walk, zhou2024efficient} explores using RL to fine-tune pretrained policies directly on physical robots, thereby bridging the gap between offline training and real-world deployment. 
SERL~\cite{luo2024serl}, an open-source software suite, supports on-robot RL through classifier-based rewards and human-labeled sparse signals. 
DPPO~\cite{dppo2024}, which fine-tunes diffusion-based policies on real hardware, demonstrates the potential of combining generative policy architectures with on-robot optimization.
In this work, we integrate semi-sparse human reward labels with dense tactile-based rewards. This hybrid formulation leverages the efficiency of human guidance while grounding the learning process in rich contact information, enabling more reliable improvement of base policies during real-world fine-tuning.

%% file: text/design.tex
\section{Sensor Design}
\label{sec:design}

SpikeATac is a multimodal tactile fingertip. Its core feature is a 16-taxel PVDF array which provides very high sensitivity to high frequency events, like making and breaking contact, extrinsic contact, and vibrations from surface textures. The PVDF is embedded near the surface of the finger's elastomer (Fig.~\ref{fig:candy}) in order to be sensitive to these types of signals while being isolated from structural vibration. This design allows us to complement the dynamic PVDF by embedding a number of commercial off-the-shelf (COTS) capacitive taxels deeper under the surface in order to provide a static pressure response. Lastly, a 3-axis accelerometer acts as a secondary dynamic modality and is connected to a ``nail" feature for surface exploration; however, we do not make use of this feature in this study.

PVDF produces charge proportional to applied strain.
We use charge amplifiers to measure the charge generated at each electrode, therefore measuring applied strain. 
However, the feedback resistor of the charge amplifier provides a discharge path, creating high-pass filter behavior.
A critical part of our work is that this configuration allows the PVDF to be very sensitive to transients without permanently saturating.
See Section \ref{sec:design.electronics} for more electronics details.

The total material cost is US\$365, dominated by the COTS capacitive pads (US\$263 for seven such sensors alone).

\subsection{PVDF Fabrication}

The PVDF stack is a flexible, transparent piezoelectric film with metal on both sides forming 16 capacitor units.
The taxelized side, fabricated using photolithography, contains pads, traces, and alignment marks. The common ground side, fabricated using a shadow mask, contains a single pad.

Each substrate is \SI{4.5}{\centi\meter} x \SI{4.5}{\centi\meter} x \SI{100}{\micro\meter} thick pre-poled PVDF (PolyK).
We sonicate the PVDF substrates for 15 min. in isopropyl alcohol and then rinse in deionized water to clean the surface before fabrication. 
A metal stack of \SI{20}{\nano\meter} of a chromium adhesion layer and \SI{200}{\nano\meter} of gold is deposited in a physical vapor deposition (PVD) system (Angstrom) with electron-beam evaporation.
Then, we spin-coat positive photoresist (S1805, Dow) onto the sample. Using a photomask written with laser lithography (Heidelberg Systems), the photoresist is patterned by a mask aligner system (SÜSS). 
The photoresist pattern is developed in AZ 300 metal ion free developer, then transferred to the chrome/gold stack with two wet etches (Transene).
This completes the taxelized side of the sample.
We deposit the common ground plane in a PVD system through a shadowmask, with the same metal stack as the taxelized side. 

To prepare the samples for SpikeATac fabrication, 
we cut the samples to the desired shape using alignment marks in the pattern. We use copper tape and a low-temperature curing silver paste (Micromax) to short the ground plane to a ground trace on the taxelized side which goes to the connector.
Finally, the sensors are hot bar bonded to an flexible printed circuit (FPC) cable using an anisotropic conductive film (Hitachi) as the conductive interface. The FPC connects to a port on the printed circuit board (PCB) described in the next section.

\begin{figure}
\centerline{\includegraphics[width=\linewidth]{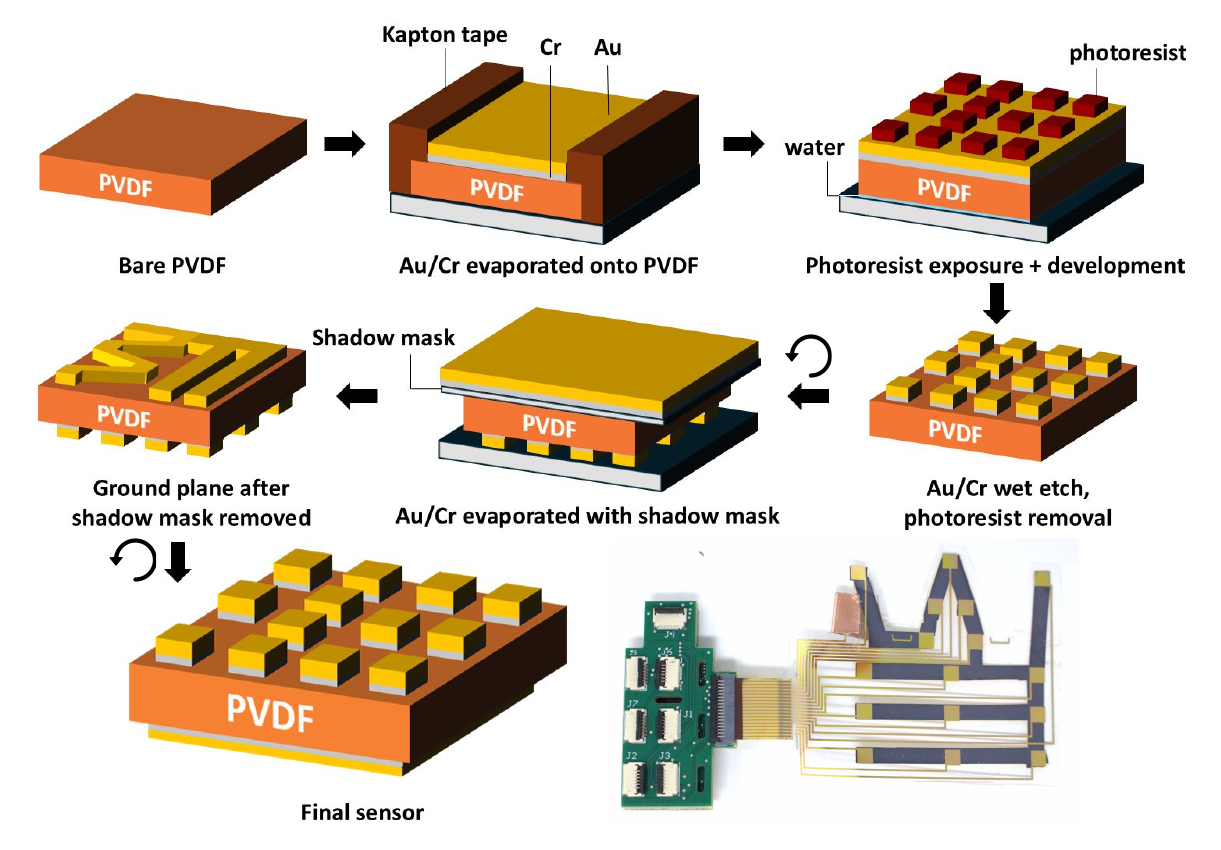}}
\caption{The PVDF fabrication process along with a photo of the finished sensor attached to the PCB stack with an FPC cable.}
\label{fig:fab}
\end{figure}

\subsection{Electronics Design}
\label{sec:design.electronics}
The ``on-finger" electronics consist of a 2-PCB stack housed within the finger. The PCBs are connected with board-to-board connectors (Molex SlimStack) and handle all amplification and sampling. Flat flexible cable (FFC) ports interface with the capacitive sensor tails and the PVDF's FPC cable. A capacitance-to-digital converter (CDC, AD7147) samples the capacitive sensors. The stack also contains the three-axis accelerometer (MC3479).

PVDF’s high impedance output requires careful electronics for sensitive, low-noise signals.
Each of 16 PVDF signals are passed through a charge amplifier ($R_f = 1.2 G\Omega$, $C_f = 22 pF$) using a high impedance JFET op amp (AD8643), then sampled with an analog-to-digital converter (ADC, ADS7953), biased to \SI{2.5}{\volt}. Our configuration provides a charge to voltage gain of $-\frac{1}{C_f} = -\frac{10^{12}}{22}$, high-pass filter cutoff frequency of $\frac{1}{2\pi R_f C_f} =$ \SI{6.0}{\hertz}, and a time constant of $R_fC_f =$ \SI{30}{\ms}. 

The on-finger PCB stack is connected to an ``off-finger" Teensy 4.1-based control board and communicates using serial peripheral interface (SPI). The control board controls up to 4 SpikeATacs and 4 Dynamixel chains and is what we use in all of our experiments. The Teensy communicates with a personal computer (PC) over USB using micro-ROS. When running only SpikeATacs, the PVDF is sampled at \SI{4}{\kilo\hertz}, the accelerometer at \SI{1}{\kilo\hertz}, and the capacitive sensors at \SI{40}{\hertz}.

\subsection{Complete SpikeATac Fabrication}

Finger fabrication contains 4 steps: mold preparation, skeleton preparation, casting, and PVDF layer application.

\subsubsection{Mold Preparation}
Half of a two-part mold (Clear V5 resin, Formlabs; washed and 15 min./\SI{60}{\celsius} UV cure) controls the front finger surface and contains a feature to create a \SI{1.5}{\mm} indentation on the finger surface for the PVDF.  
Once per mold, we use a brush to apply 2 layers of Inhibit X (Smooth-On) to prevent cure inhibition \cite{venzac_pdms_2021} (outgassing each coat for 30 min). The other half of the mold (PLA) controls the back of the finger geometry. At every use, we spray a thin coating of release agent (EaseRelease200) onto both molds.

\subsubsection{Skeleton Preparation}
A rigid skeleton forms the structure of the finger (Grey V5 resin, Formlabs; washed and 15 min./\SI{60}{\celsius} UV cure).
After inserting pocket nuts, we adhere 7 capacitive sensors (Singletact, \SI{8}{\mm}/\SI{10}{\newton} model, PPS) onto the skeleton using double-sided tape (300LSE, 3M). We route the Singletacts' \SI{50}{\mm} tails into the skeleton's internal space and connect them to the PCB's FFC ports. We apply 2 coats of Inhibit X to the skeleton (outgassing each coat for 30 min.), followed by 1 coat of silicone primer (SS4120, Momentive) to encourage bonding.

\subsubsection{Casting}
We attach a JST cable, carrying power and SPI buses to the control board, to the on-finger PCB. We suspend the skeleton in the mold by fastening the PLA negative mold to the back of the skeleton (M1.6 screws), which also fixes the PCB stack in its permanent position. We assemble the mold with screws and an O-ring. We mix and degas Ecoflex 00-50 elastomer (Smooth-On) and pour it into the mold, allowing it to cure at room temperature for \SI{8}{\hour}.

\subsubsection{PVDF Layer Application}
Applying PVDF is nontrivial due to its imperfect fit on the finger's multi-curved surface and the difficulty of achieving strong PVDF/elastomer adhesion.
We first adhere the PVDF to a flat, \SI{0.5}{\mm} thick Ecoflex 00-50 elastomer sheet of the same shape (Sil-Poxy, Smooth-On). After curing, we adhere this elastomer/PVDF stack into the finger surface's indentation (Sil-Poxy). We find that this 2 step process provides some forgiveness in the alignment to the finger surface and provides improved adhesion as the problematic elastomer/PVDF interface is cured completely while flat. Finally, we paint thin layers of Ecoflex 00-50 on top of the PVDF to form a thin layer ($\sim$3 coats).

%% file: text/characterization.tex
\section{Characterization and Sensitivity Analysis}
\label{sec:characterization}


\begin{figure}
\centerline{\includegraphics[width=\linewidth]{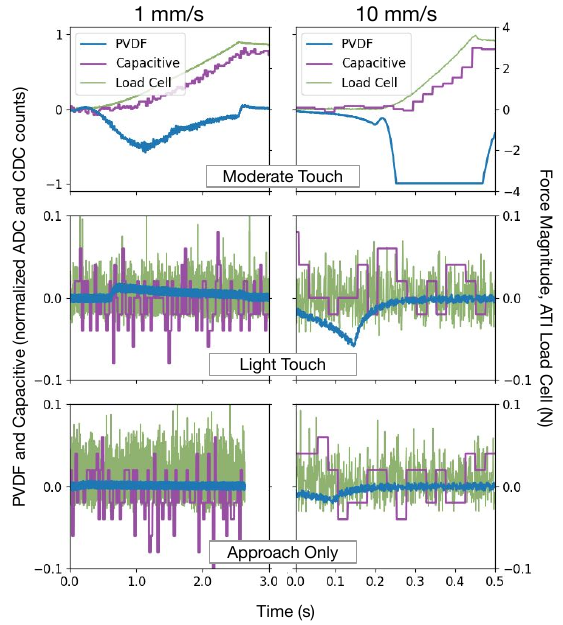}}
\caption{PVDF, capacitive, and ATI Gamma responses when probing with a hemispherical indenter ($d=\SI{6}{mm}$, see Fig.~\ref{fig:heatmap}) at two linear probe speeds: \SI{1}{mm/s} (left) and \SI{10}{mm/s} (right). We test when under a moderate touch condition ($\sim$\SI{3.5}{\newton}), light touch condition (minimal contact established visually), and when approaching but not touching. Bottom-center PVDF and capacitive taxels are shown (location 8 on Fig.~\ref{fig:heatmap}). ADC counts are normalized to the full measurement range; CDC counts are normalized to just above the moderate touch's maximum response. We note that, at the higher speed, PVDF can detect light contact that the other sensors cannot, or detect contact earlier than the other sensors.}
\label{fig:linear_probe}
\end{figure}

\begin{figure}
\centerline{\includegraphics[width=\linewidth]{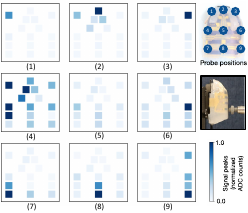}}
\caption{Heatmaps of the maximum absolute value of the zeroed PVDF response from each taxel (i.e., a dark taxel represents an increasing or decreasing signal) depending on indentation location, showing the spatial information carried by the PVDF taxels.}
\label{fig:heatmap}
\end{figure}

To characterize the signal response of the different modalities, we mount the sensor on an ATI Gamma 6-axis force/torque (F/T) load cell (SI-130-10 calibration) and probe it with a hemispherical indenter ($d=\SI{6}{mm}$) on a linear probe at \SI{1}{\mm/\s} and \SI{10}{\mm/\s}.
The finger is probed in 3 conditions: \texttt{moderate touch} ($\sim$\SI{3.5}{\newton}), \texttt{light touch} (probe just barely makes visual contact), and \texttt{approach only} (probe stops \SI{0.5}{\mm} away from finger. Sensor responses are shown in Fig. \ref{fig:linear_probe}.

We note that, at the slow speed, PVDF provides a lower response and the signal dies away quickly. In other words, at this probe velocity, the high-pass filter behavior of the charge amps dominates the response. However, for faster incoming speed, the PVDF signal is rich in information: PVDF is very sensitive to the initial contact, spiking to saturation even at very low force.
In fact, the PVDF signal starts to spike before the load cell or capacitive sensors move above their noise floor, showing that the PVDF's minimum detectable force at this probe velocity is below the load cell's noise floor (1 standard deviation of noise is \SI{20}{\mN}). The \texttt{light touch} condition also supports this conclusion, as the PVDF responds and the load cell does not.

The \texttt{approach only} data shows that the PVDF also exhibits a small response to object proximity, largely due to static charge, but this effect is small in comparison to true contact signals, even for \texttt{light touch}. We empirically observe that the proximity effect increases after adding the elastomer layer.

The taxelized nature of our PVDF sensor is a key feature of this work. To test its behavior, we also probe the finger in 9 different locations to evaluate spatial information contained in the PVDF signals. These indentations are  $\sim$\SI{1.5}{\newton} and normal to the surface (via different finger mounts). A heatmap of the maximum absolute signals from each taxel (Fig. \ref{fig:heatmap}) shows that the PVDF provides spatial contact information. We note that the traces routed on the sample also contribute signal. For example, the indentation at position 9 (bottom right) activates the upper right taxel because the upper right taxel's trace is routed near position 9 (see trace layout in Fig. \ref{fig:fab}). We note that position 5 has a weaker response due to a disconnected trace in the finger used for testing.

\section{Fast and Delicate Grasping}

\begin{figure}[t] 
\centering
\includegraphics[width=\linewidth]{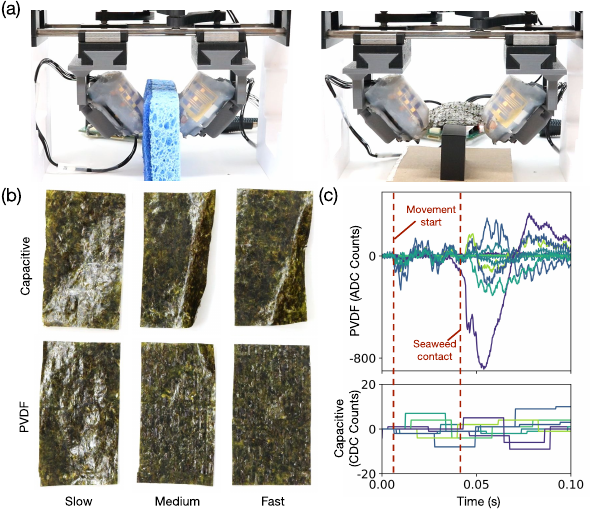}
\caption{(a) Setup of the fast and delicate grasping experiment, in which we command a \texttt{slow}, \texttt{medium}, or \texttt{fast} velocity to the gripper and stop after contact is detected, comparing PVDF-based and capacitive-based detection (Tab. \ref{tab:fastgrasp}). (b) After 15 trials, the more fragile object (seaweed) is noticeably crumpled when using capacitive-based stopping, but appears nearly untouched in the PVDF condition at \texttt{fast} velocity. In fact, in the \texttt{medium} and \texttt{fast} conditions, the capacitive method often failed to detect contact at all. (c) Raw signals from one finger show a distinct PVDF spike at seaweed contact, while the capacitive signal does not reliably rise above its noise floor (each line/color shows an individual taxel from one of the two fingers).}
\label{fig:response_time}
\end{figure}

Utilizing PVDF's ability to quickly and sensitively detect the onset of contact, we study SpikeATac's ability to react fast enough to effectively stop on impact while moving at high speed.
Moving fast while still being delicate is a valuable skill for robots moving quickly in unstable or safety critical environments and to grasp fragile objects without necessitating slow, time consuming movements. We thus compare how different sensing modalities enable this ability.

\subsection{Experiment Setup}
We mount two SpikeATacs on a custom parallel gripper containing a servo motor (XM430-210, Dynamixel) and a linkage mechanism controlling the linear motion (Fig. \ref{fig:response_time}).
We mount the gripper on a stand and place objects on a table between the fingers.
During each grasp, we send a velocity command to the Dynamixel's velocity controller (P=300, I=3500), followed by a stop command (velocity = 0) once contact is detected (see below for details on contact detection). We check for the stop condition at 4 kHz (3 kHz for \texttt{slow} velocity due to Teensy memory limits), with PVDF and capacitive data updating at 4 kHz and 40 Hz, respectively. The PVDF has a median filter of width 3.
The capacitive data is not filtered, however we remove occasional outlier values introduced by SPI bus noise.

We compare two methods of contact detection, one using PVDF, and the other using the capacitive sensors. In each case, we selected what we found to be the best performing method. For PVDF, we found a difference-based method to be the most effective, as the PVDF responds with sharp edges for the velocities investigated. However, we found this not to be the case for the capacitive signal, which does not respond as sharply, and we therefore selected a method which looks at the average signals compared to a static baseline. 
\begin{itemize}
    \item \textit{PVDF Method}: Contact occurs once $\geq2$ PVDF taxels show a current–previous difference above a threshold.
    \item \textit{Capacitive Method}: Contact occurs when the mean of the seven capacitive taxels moves beyond a threshold.
\end{itemize}
In each case, we chose thresholds empirically to be just above the noise floor and such that the motor doesn't stop prematurely due to noise at the movement start.

The methods above are only for the \textit{initial} contact detection of either finger. In both methods, to ensure sufficient grasp force for lifting the object, we follow the contact detection with a slow movement that stops when the mean of all capacitive values passes a second, object-specific threshold. For the PVDF and capacitive methods, respectively, we use a threshold of (40, 80) and (5.5, 6.5) counts for (finger 1, finger 2). We use a capacitive threshold of 3 (seaweed) and 5 (sponge) counts for the stable contact adjustment.

To run the experiment, we command one of three velocities to the motor and grasp one of two experimental objects: a sponge or seaweed (nori) sheets. Once contact is detected, the measured motor velocity at that timestep is recorded and the stop command is sent, followed by adjustment with capacitive feedback. We then record how far the gripper traveled after the ground truth contact point (inferred from object width), as well as the linear velocity of the gripper near the contact.

We run each of 12 experiments (two objects, two methods, three velocities) for 30 trials, verifying that the object is in a stable grasp at the end of each trial. Commanded velocities are the same for the two objects, however due to different object width, actual average achieved velocities were (91, 90) (\texttt{slow}), (180, 172) (\texttt{medium}), and (281, 180) mm/s (\texttt{fast}) for the (sponge, seaweed).
We replace and remeasure the seaweed for each of six conditions.

\subsection{Results}

\begin{table}[t!]
\renewcommand{\arraystretch}{1.5}
\caption{Distance traveled by the gripper (mean and stdev) beyond start of contact (smaller distance = faster stop) over 30 trials as illustrated in Fig. \ref{fig:response_time}. We find that PVDF enables faster stopping, especially for the fragile object (seaweed), whereas the capacitive sensors are not able to reliably stop before crushing the object in the medium and fast conditions.}
\label{tab:fastgrasp}
\centering
\footnotesize
\setlength{\tabcolsep}{3pt}  
\begin{tabular}{c|ccc|ccc}
                                & \multicolumn{3}{c|}{Sponge}
                                & \multicolumn{3}{c}{Seaweed}
                                \\ \hline \hline
                                Velocity & \multicolumn{1}{c|}{Slow} & \multicolumn{1}{c|}{Medium} & Fast &
                                \multicolumn{1}{c|}{Slow} & \multicolumn{1}{c|}{Medium} & Fast
                                \\ \hline
                                \multicolumn{1}{c|}{Cap. (mm)} &
                                \multicolumn{1}{c|}{\textbf{0.8$\pm$0.4}} & \multicolumn{1}{c|}{1.9$\pm$0.7} & 3.7$\pm$0.5 &
                                \multicolumn{1}{c|}{13.3$\pm$2.3} & \multicolumn{1}{c|}{--}  & --
                                \\ \hline
                                \multicolumn{1}{c|}{PVDF (mm)} & 
                                \multicolumn{1}{c|}{1.2$\pm$0.2} & \multicolumn{1}{c|}{\textbf{1.2$\pm$0.4}} & \textbf{1.7$\pm$0.1} &
                                \multicolumn{1}{c|}{\textbf{3.6$\pm$3.8}} & \multicolumn{1}{c|}{\textbf{1.9$\pm$0.6}} & \textbf{2.4$\pm$0.4}    
\end{tabular}
\end{table}

The sponge provides a quantitative comparison between the two modalities as it is deformable but firm enough that both methods can reliably grasp it gently (Tab. \ref{tab:fastgrasp}).
While there is little difference between methods at the \texttt{slow} velocity, the difference at higher velocity is apparent and consistent. PVDF allows the gripper to respond faster upon initial contact at the high velocities than when using capacitive alone. These results are consistent with the expected behavior given that PVDF's response is faster and more sensitive.

The results on the more fragile object (seaweed) are more striking. Over 30 trials at each speed, grasping with the PVDF sensors succeeded every time (0/30 crushed objects at each speed). In contrast, grasping with capacitive sensing succeeded at \texttt{slow} speeds (0/30 crushed objects) but failed often at \texttt{medium} and \texttt{fast} speeds (20/30 and 23/30 crushed objects, respectively). Additionally, the seaweed sheet after the high velocity PVDF condition appears nearly untouched. Due to the PVDF's high pass filter behavior and detection algorithm, the faster we move with PVDF the more sensitive it will be. An illustration of the seaweed sheets after 15 trials is provided in Fig. \ref{fig:response_time}.

We note that we cannot rule out proximity signals caused by static charge contributing to the PVDF's contact detection.
However, signals caused by proximity tend to be small relative to contact signals (Fig. \ref{fig:linear_probe}), at least for the objects investigated in these experiments. The sharp edge and large magnitude in the raw PVDF data (Fig.~\ref{fig:response_time}) and consistent stopping behavior suggest the sensing of the contact event itself plays a critical role.

Finally, we use the fast grasping capability demonstrated here to grasp and pick up a wide set of delicate objects. See the \href{https://roamlab.github.io/spikeatac/}{supplementary video} for examples.

%% file: text/rh3.tex
\section{Learning Delicate Object In-hand Rotation}
\label{sec:rh3}

So far, we have focused on standalone characterization of SpikeATac, as well as its integration with a parallel gripper for fast yet delicate manipulation. However, a traditional limitation of complex tactile sensors is the difficult integration in fully dexterous, multifingered manipulation. This is particularly true for current data-driven sensorimotor policies, for example for in-hand object re-orientation, which often train in simulation~\cite{khandate_r_2024, wang_lessons_2024, qi_general_2023}: if a sensor provides highly complex signals which are difficult to simulate, how can it be used for learned motor control?

\begin{figure}[t]
    \centering
    \includegraphics[width=\linewidth]{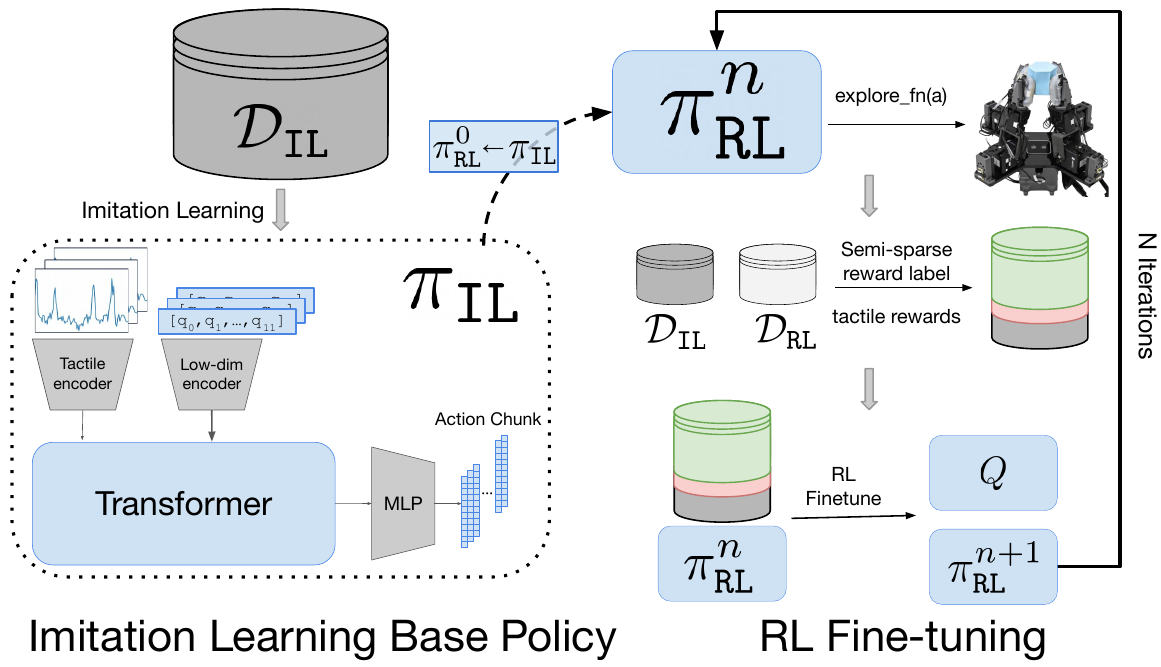}
    \caption{Policy learning pipeline using on-robot RLHF in conjunction with imitation learning to fine-tune a control policy using real sensor data.}
    \label{fig:ilrl}
\end{figure}

To address this challenge, we integrate SpikeATac into a dexterous, four-finger robot hand, and use a reinforcement learning from human feedback (RLHF) pipeline (Fig.~\ref{fig:ilrl}) that is designed for fine-tuning behaviors on the real robot using raw sensor signals. To demonstrate SpikeATac's sensitive and multimodal capabilities, we also tackle the problem of in-hand rotation via finger gaiting with only tactile and proprioceptive information. However, we make the problem even more difficult by extending it to fragile objects. To the best of our knowledge, this ability has not been previously shown in the literature.

\subsection{On-Robot RL Fine-tuning for Behavior Improvement}
At a high level, we start with a base policy $\pi_{\mathtt{IL}}$ that operates on raw sensor signals from our robot. To obtain $\pi_{\mathtt{IL}}$, we train a behavioral policy in simulation via RL using a highly simplified, binary contact signal (touch vs. no-touch). We transfer this policy to the real robot using standard domain-randomization, and use it to collect a number of demonstrations on our real platform, which are in turn used to train a base policy $\pi_{\mathtt{IL}}$ via imitation learning. 

While $\pi_{\mathtt{IL}}$ works well on hard objects (shown in our \href{https://roamlab.github.io/spikeatac/}{supplementary video}), it immediately fails on fragile ones, as it has been trained with rigid objects and lacks the ability to produce delicate touch. We thus fine-tune it using RL, specifically Soft Actor-Critic~\cite{haarnoja2018soft}, on raw sensor data acquired from our sensor. This pipeline shows how the highly-sensitive, difficult-to-simulate sensor signals can be integrated into learning-based manipulation. 

To initialize RL fine-tuning, we set $\pi^0_{\mathtt{RL}} = \pi_{\mathtt{IL}}$ and deploy it on the real robot to collect data. To enable effective data collection for training the Q-function, we introduce an exploration strategy by injecting Gaussian noise with small standard deviations into the distal joints. At each iteration, we use the most recent policy $\pi^n_{\mathtt{RL}}$ to collect a dataset $\mathcal{D}^{n}_{\mathtt{RL}}$. For training, we aggregate this RL dataset with the original imitation learning dataset, i.e., $\mathcal{D}_{\mathtt{IL}} \cup \mathcal{D}^n_{\mathtt{RL}}$.

For RL fine-tuning, we use a task-specific reward $r_{\mathtt{task}}$ provided by human annotators.\footnote[2]{An earlier version of this paper incorrectly stated that we also use a tactile component in the reward function. In fact, the version of the policy shown here does not make use of explicit tactile rewards. However, it still optimizes behavior based on real-world tactile data, and incorporates tactile information when fine-tuning through the human-provided labels.} For each trajectory, while viewing a playback of camera and tactile data, a human labels segments as ``good” if the object is rotating and ``bad” otherwise. We refer to this as semi-sparse reward labeling. Compared to sparse rewards, where only the final timestep is labeled based on whether the agent achieves the goal, this approach provides richer feedback across different parts of a trajectory. At the same time, unlike dense rewards, our reward labels do not require overly complex state estimation or engineering in the real-world setup.

\subsection{Observation and Action Space}
We carefully handle raw PVDF sensor signals in the observation because we do not want to discard high frequency information in the sensor data. We sample the PVDF at \SI{450}{\hertz} and capacitive sensors at \SI{250}{\hertz} (capacitive internal update rate is still \SI{40}{\hertz}), while the policy runs at \SI{20}{\hertz} (sampling frequencies are limited by the bandwidth of the microcontroller controlling the entire hand). The signals have a median filter of width 5. A history buffer of 64 PVDF and capacitive readings incorporates higher frequency information into the policy at each lower frequency timestep.

The observation contains the length 64 history buffer of sensor signals (16 PVDF and 7 capacitive signals per finger) and the 12 joint encoder readings. The action space is the hand's 12 joint position set points for its 12 degrees of freedom (1 roll joint and 2 flexion/extension joints per finger).

\begin{figure}
    \centering
    \includegraphics[width=0.98\linewidth]{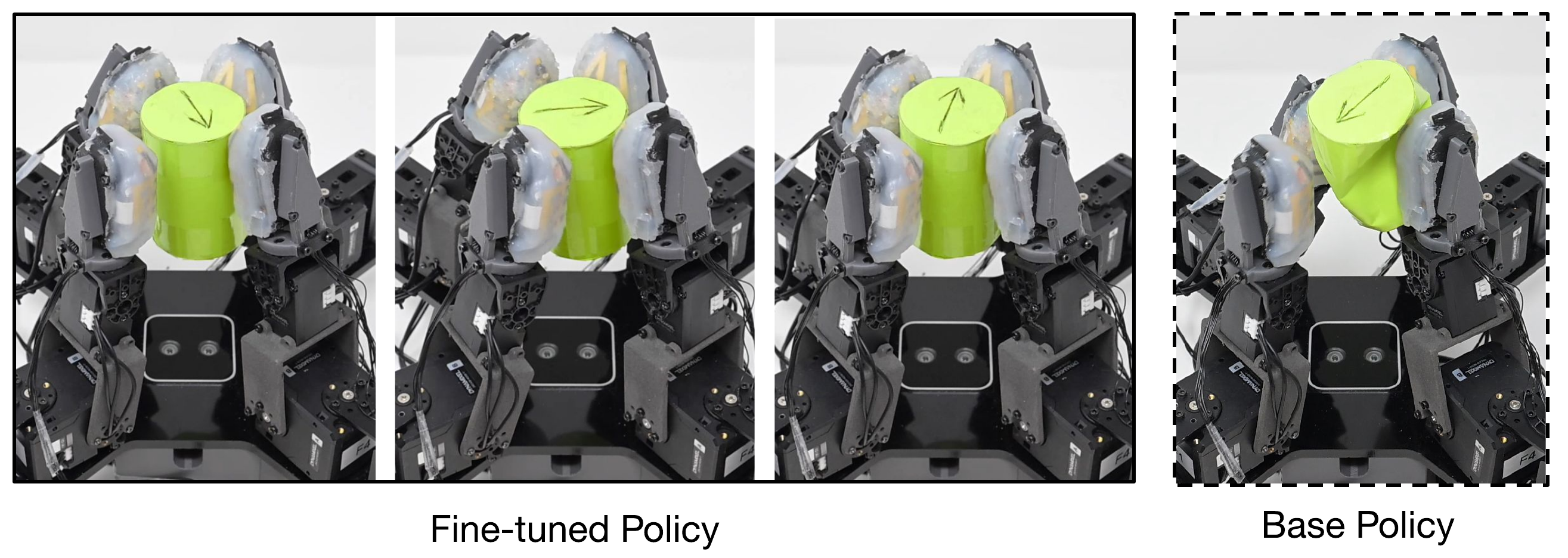}
    \caption{Rollout of the fine-tuned policy at three time points. The base policy $\pi^0_{\mathtt{RL}} = \pi_{\mathtt{IL}}$ quickly crushes the paper object, while the fine-tuned policy using raw SpikeATac signals is capable of long rollouts without damaging the object.}
    \label{fig:traj}
\end{figure}

\subsection{Results}
\begin{figure}
    \centering
    \includegraphics[width=0.9\linewidth]{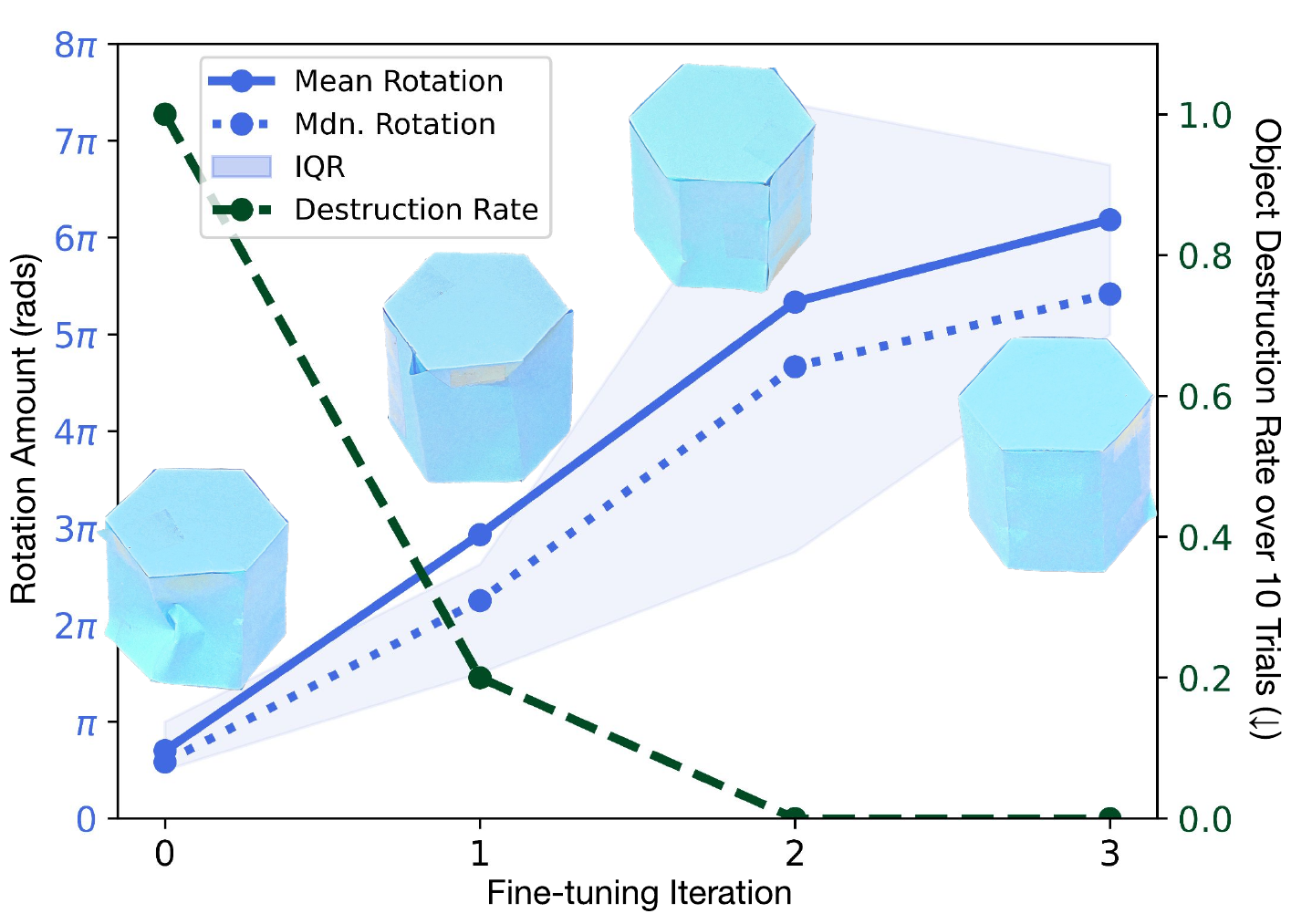}
    \caption{Mean and median rotation amount and object destruction rate (\# times crushed out of 10) with each RL fine-tuning learning iteration evaluated on a paper hexagonal prism and cylinder (figure shows combined results). The images show examples of objects after manipulation at each iteration. The shaded region shows the interquartile range (IQR).}
    \label{fig:rh3_results}
\end{figure}

To evaluate performance, we roll out the policy at each iteration on a paper hexagonal prism (d = h = \SI{40}{\mm}) and paper cylinder (d = \SI{40}{\mm}, h = \SI{60}{\mm}) for 5 rollouts each. We evaluate amount of rotation and destruction rate for each policy (Fig. \ref{fig:rh3_results}).

The initial $\pi^0_{\mathtt{RL}} = \pi_{\mathtt{IL}}$ policy almost immediately crushes the objects (Fig. \ref{fig:traj}), however through this learning pipeline the policy learns to make visibly softer contacts over the learning iterations, as evidenced by the object state after the experiments (Fig. \ref{fig:rh3_results}), while amount of rotation also increases. See the \href{https://roamlab.github.io/spikeatac/}{supplementary video} for visual results of improved rotation over learning iterations.

We compare RL fine-tuning results to an IL baseline that is trained on all available demonstration data after all fine-tuning iterations, with the IL baseline achieving 5.7/2.9 (mean/mdn.) radians rotation and 0.7 destruction rate, notably lower than the RL fine-tuning results. We emphasize that the data we use for the IL fine-tuning baseline is usually hard to obtain in an IL setting, given the quick failures of the base policy. Here, we filter the RL fine-tuning dataset using human labels and only use the good trajectories for imitation learning. 

These experiments show that this system and pipeline as a whole enable this difficult, delicate task.
We note that the resulting policy in these experiments may be overfit to these particular objects and object sizes, and that we also don't have a definitive analysis of what each sensor contributes.
Our primary takeaway, however, is that this pipeline, which is designed for a real robot using real sensor signals in the reward, allows the use of SpikeATac data for new data-driven capabilities, despite the sensor being, as of now, very difficult to simulate. 

%% file: text/discussion.tex
\section{Discussion and Conclusions}
\label{sec:discussion}

Relative to other tactile sensing transducers, PVDF is a less prevalent choice of tactile modality in the literature despite its impressive sensitivity. This may in part be due to its susceptibility to noise, as a very high impedance system, and its cumbersome fabrication processes--the clean room tools used in this work are not readily accessible to many robotics-focused researchers. 

A differential amplifier design and more advanced shielding will likely improve the PVDF's performance \cite{yeiser_umbomic_2024}. To improve connector robustness, we'd also like to eliminate the ACF by connecting PVDF electrodes directly to a flexPCB \cite{yeiser_umbomic_2024}. We note that in addition to sensitivity to static charge, our sensor is also sensitive to how the system is grounded, but we found that with proper grounding the sensor signals were robust enough to enable all the experiments presented here.

Additionally, although the clean room processes used in this work make ``democratizing" this sensor for the research community challenging, they are standard fabrication processes and should not be feared at scale. Screen printing may also be another avenue for scalable fabrication. In this vein, with careful implementation, we see PVDF as a compelling tool for many-taxel dynamic tactile sensing.

%% file: text/conclusion.tex
Overall, we believe SpikeATac is a powerful platform for studying multimodal tactile sensing and manipulation. The sensor's many-taxel PVDF enables fast and delicate grasping. In conjunction with on-robot RL fine-tuning, we can integrate its raw, difficult-to-simulate multimodal sensor signals into complex learning-based policies that show unprecedented capabilities, such as in-hand reorientation of fragile objects. We believe that the combination of static and dynamic tactile sensing, integrated into capable hands and used in conjunction with powerful sensorimotor learning methods, shows great promise for future advances in robot dexterity.

%% file: text/ack.tex
\section{Acknowledgements}
\label{sec:ack}
We thank Trey Smith, Brian Coltin, Amr El-Azizi, and Rajinder Singh Deol for insightful discussion.